\pdfoutput=1

\documentclass[11pt]{article}

\usepackage{acl}

\usepackage{tikz}
\usepackage{pgfplots}
\usepackage{pgfplotstable}
\pgfplotsset{compat=1.18}
\usetikzlibrary{shapes, arrows, positioning, pgfplots.groupplots}

\usepackage{times}
\usepackage{latexsym}
\usepackage{hyperref}
\usepackage{listings}
\usepackage{booktabs}
\usepackage{tabularx}
\usepackage{multirow}
\usepackage{boldline}
\usepackage{subcaption}

\usepackage{siunitx}

\usepackage[T1]{fontenc}

\usepackage[utf8]{inputenc}

\usepackage{microtype}

\usepackage{inconsolata}

\usepackage{graphicx}

\title{Generating bilingual example sentences with large language models as lexicography assistants}

\author{Raphael Merx \hspace{1cm} Ekaterina Vylomova \hspace{1cm} Kemal Kurniawan \\
        School of Computing and Information Systems, The University of Melbourne \\
        \texttt{rmerx@student.unimelb.edu.au}\\
        \texttt{\{vylomovae, kurniawan.k\}@unimelb.edu.au}
        }

\robustify\bfseries

\begin{document}
\maketitle
\begin{abstract}

We present a study of LLMs' performance in generating and rating example sentences for bilingual dictionaries across languages with varying resource levels: French~(high-resource), Indonesian~(mid-resource), and Tetun~(low-resource), with English as the target language. We evaluate the quality of LLM-generated examples against the GDEX~(Good Dictionary EXample) criteria: typicality, informativeness, and intelligibility~\citep{kilgarriff2008gdex}. Our findings reveal that while LLMs can generate reasonably good dictionary examples, their performance degrades significantly for lower-resourced languages. We also observe high variability in human preferences for example quality, reflected in low inter-annotator agreement rates. To address this, we demonstrate that in-context learning can successfully align LLMs with individual annotator preferences. Additionally, we explore the use of pre-trained language models for automated rating of examples, finding that sentence perplexity serves as a good proxy for "typicality" and "intelligibility" in higher-resourced languages. Our study also contributes a novel dataset of 600 ratings for LLM-generated sentence pairs, and provides insights into the potential of LLMs in reducing the cost of lexicographic work, particularly for low-resource languages.

\end{abstract}


\section{Introduction}

Example sentences in bilingual dictionaries play a crucial role in language learning, helping L2 speakers to understand the meaning of headwords (words that mark a separate entry in the dictionary), and their usage in context \cite{potgieter2012example, nielsen2014example, caballero2024theory}. What makes candidate sentences good as examples is the subject of linguistic research, with \citet{kilgarriff2008gdex} proposing the GDEX (Good Dictionary EXample) framework, which qualifies good examples as typical ("exhibiting frequent and well-dispersed patterns of usage"), intelligible ("avoiding gratuitously difficult lexis and structures"), and informative ("helping to elucidate the definition"), as illustrated in Table~\ref{tab:gdex-illustration}. In bilingual setups, the accuracy of translation between source and target examples also contributes to example quality.

\begin{table}[!t]
\centering
\small
\begin{tabularx}{\linewidth}{X}
\toprule
\textbf{Typical}: Show how the word is commonly used. \newline 
\colorbox{green!15}{Yes} The business was highly \underline{successful}, turning a profit in its first year. \newline
\colorbox{red!15}{No} The \underline{successful} completion of his puzzle took months. \\
\midrule
\textbf{Informative}: Provide additional clarity beyond the word definition. \newline 
\colorbox{green!15}{Yes} Her marketing campaign was \underline{successful}, resulting in a 50\% increase in sales. \newline
\colorbox{red!15}{No} They were \underline{successful}. \\
\midrule
\textbf{Intelligible}: Easy to understand, not overly complex. \newline 
\colorbox{green!15}{Yes} The students were \underline{successful} in completing their group project on time. \newline
\colorbox{red!15}{No} Notwithstanding the exigencies of the situation, the team's herculean efforts proved \underline{successful}. \\
\bottomrule
\end{tabularx}
\caption{GDEX criteria definitions and English example sentences for the word "successful", with one sentence that fulfils the criterion and one that does not.}
\label{tab:gdex-illustration}
\end{table}

The extensive work required to come up with example sentences increases the cost of compiling lexicographic resources \cite{he_controllable_2022}. This has prompted research into the automatic selection of example sentences from existing corpora~\citep{kilgarriff2008gdex, frankenberg2014use}. However, existing corpora might not always contain sentences that are suited to language learning, as their text can be overly complex, fail to further explain the meaning of the headword, or not be licensed for reproduction. As a result, researchers have begun exploring models tailored for the generation of dictionary example sentences from a headword and its dictionary definition \cite{he_controllable_2022}.

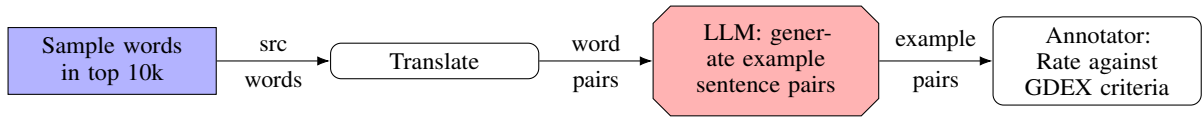
\begin{figure*}[!ht]
\centering{
\resizebox{\textwidth}{!}{\usetikzlibrary{shapes.geometric, arrows.meta, positioning}

\tikzset{
    base/.style = {draw, text width=2.5cm, align=center, minimum height=1em, font=\small},
    inout/.style = {base, rectangle, rounded corners},
    process/.style = {base, rectangle, fill=blue!30},
    gate/.style = {base, chamfered rectangle, fill=red!30},
}

\begin{tikzpicture}[node distance=1.2cm and 0.5cm, auto]
    \node [process] (top50) {Sample words in top 10k};
    \node [inout, right=1.5cm of top50] (translate) {Translate};
    \node [gate, right=1.5cm of translate] (llm) {LLM: generate example sentence pairs};
    \node [inout, right=1.5cm of llm] (rate) {Annotator: Rate against GDEX criteria};

    \draw[-Latex] (top50) -- (translate) 
        node[midway,above] {\small src } 
        node[midway,below] {\small words};
        
    \draw[-Latex] (translate) -- (llm) 
        node[midway,above] {\small word } 
        node[midway,below] {\small pairs};
        
    \draw[-Latex] (llm) -- (rate) 
        node[midway,above] {\small example } 
        node[midway,below] {\small pairs};

\end{tikzpicture}}}
\caption{Overview of our process for generating example sentence pairs using LLMs.}
\label{fig:process-generate-examples}
\end{figure*}

Large language models (LLMs) trained on a wide range of texts \cite{Gao2020ThePA} might be well suited to formulate generic and informative example sentences that benefit language learning. In particular, their capacity to adapt to new, unseen tasks \cite{Radford2019LanguageMA, kojima2023largelanguagemodelszeroshot} means that they might be well suited to generate sentences against specific criteria. However, questions about the quality of the sentences they generate, and their ability to understand what makes a good example, remain.

In this paper, we review LLMs capability to generate and rate example sentences in a bilingual lexicography context, against the GDEX criteria. We work with three language pairs, with English on the target side, and source sides that cover a range of language resource levels: French (high-resource), Indonesian (mid-resource), Tetun (low-resource). The paper makes the following contributions:
\begin{itemize}
    \item An evaluation of LLMs capability to \emph{generate} bilingual example sentence pairs, across languages of different resource levels;
    \item An evaluation of pre-trained models and LLMs capability to \emph{rate} the generated bilingual example pairs, both against the GDEX criteria (qualitative), and against an overall rating (quantitative, 1-5);
    \item A novel dataset of 600 sentence ratings for LLM-generated example sentence pairs in French, Indonesian, and Tetun as source, and English as target. Each pair is rated against 5 criteria, resulting in 3,000 individual annotations.\footnote{\href{https://github.com/raphaelmerx/llm-bilingual-examples}{https://github.com/raphaelmerx/llm-bilingual-examples}}
\end{itemize}

\section{Background}

\paragraph{LLMs for synthetic data generation.}
While hallucinations can make LLMs unreliable for tasks that require factual accuracy \cite{azamfirei2023large}, the text they generate can be of high quality, in some cases preferred over human-generated text by human annotators \cite{west2023generativeaiparadoxwhat, almeman-etal-2024-wordnet, cai-etal-2024-low}. LLM generation of synthetic data has several downstream applications, including the creation of corpora for subsequent training of specialised models \cite{li2023syntheticdatagenerationlarge, whitehouse2023llm} and the generation of examples to aid learning \cite{jury2024evaluating, nam2024using}. In lower resource scenarios, LLMs exhibit an increased tendency to generate inaccurate or poor quality information \cite{cahyawijaya-etal-2024-llms, benkirane2024machinetranslationhallucinationdetection}. However, this limitation is not entirely prohibitive; recent research has demonstrated that LLMs can be leveraged to generate synthetic resources when authentic materials are scarce \cite{santoso-etal-2024-pushing}. This dual nature of LLMs in low-resource contexts---their proneness to hallucination and their potential for synthetic data generation---presents both challenges and opportunities for their application in bilingual lexicography.

\paragraph{Automated extraction and generation of dictionary examples.}
The identification, rating, and generation of dictionary examples has been the subject of previous research. Using the GDEX criteria, \citet{almeman2022putting} found that many WordNet examples \cite{miller1995wordnet} are of poor quality, often because they are too short, in comparison with those from the Oxford English Dictionary \citeyearpar{dictionary1989oxford}. A subsequent study found that ChatGPT-generated examples are rated higher by human annotators than those from the Oxford Dictionary~\citep{almeman-etal-2024-wordnet}. \citet{cai-etal-2024-low} further introduced OxfordEval, an evaluation metric defined as the win rate between generated sentences and the Oxford Dictionary, and found that LLM-generated examples have over 80\% win rate. They also introduced the selection of candidate sentences through a masked language model to marginally improve the win rate. In non-English settings, results were found to be more mixed: working with Japanese, \citet{benedetti2024automatically} found human examples were still preferred by annotators, with high rates of disagreement between annotators about example quality. In a low-resource setting, working with Singlish, \citet{chow-etal-2024-word} found that ChatGPT could be leveraged to produce draft dictionary entries, including example sentences, but authors did not rate the examples independently of generated definitions.

\setlength{\tabcolsep}{4pt}
\begin{table*}
\centering
\small
\begin{tabularx}{\linewidth}{lllXXXl}
\toprule
Lang & Src & Tgt & Src sentence & Tgt sentence & GDEX ratings & Overall rating \\
\midrule
\texttt{tdt} & rai & country & Timor-Leste mak rai ida ne'ebe iha laran kultura barak. & Timor-Leste is a country rich in culture. & Typical: Yes\newline Informative: Yes\newline Intelligible: Yes\newline Transl. correct: No & 3 - Average \\
\midrule
\texttt{ind} & meriam & cannon & Meriam itu ditempatkan di atas bukit untuk melindungi kota dari serangan musuh. & The cannon was placed on the hill to protect the city from enemy attacks. & Typical: Yes\newline Informative: Somewhat\newline Intelligible: Yes\newline Transl. correct: Yes & 4 - Good \\
\midrule
\texttt{fra} & on & we & On va au cinéma ce soir. & We are going to the cinema tonight. & Typical: Yes\newline Informative: Yes\newline Intelligible: Yes\newline Transl. correct: Yes & 5 - Very good \\
\bottomrule
\end{tabularx}
\caption{Example LLM-generated sentences and annotator ratings for languages covered in this study.}
\label{tab:example-sentences}
\end{table*}

\paragraph{Research gap.} Despite the growing body of research on LLMs in lexicography, several areas remain unexplored. First, there has been no structured evaluation of LLM capabilities in generating example sentences for bilingual dictionaries, where additional challenges arise compared to monolingual dictionaries, such as maintaining GDEX criteria across languages while ensuring translation accuracy. Second, the potential of LLMs to help assess the quality of examples in a bilingual context, which could assist with example selection and with the setup of self-improvement pipelines for generation, has not been systematically investigated. Lastly, we have not found comprehensive studies examining LLM-based optimisation techniques---such as prompt engineering, fine-tuning, and in-context learning---for the specific task of generating dictionary examples. Addressing these research gaps could advance our understanding of how to effectively harness LLMs for creating high-quality, contextually appropriate example sentences in bilingual dictionaries, across languages of varying resource levels.


\section{LLM generation of bilingual example sentences}

This section describes our methodology for generating bilingual example sentences using LLMs, and results from human annotation of these generated sentences.



\subsection{Methodology for generation}

Figure~\ref{fig:process-generate-examples} provides an overview of our proposed methodology for generating and rating examples.

\paragraph{Word selection}
For each source language~(French, Indonesian, Tetun), we randomly select 50 words from the top 10,000 most frequent words. We use existing word lists for French\footnote{\href{http://www.lexique.org/}{http://www.lexique.org/}} and Indonesian,\footnote{\href{https://github.com/hermitdave/FrequencyWords/blob/master/content/2018/id/id_full.txt}{FrequencyWords/id\_full.txt}} and generate that list for Tetun by finding the top 10,000 words in the Labadain 30k dataset \cite{de_jesus_labadain-30k_2024}, the largest available Tetun dataset audited by native speakers. We then manually translate each of the 50 words to their English equivalent. When words have multiple translations, we select the one that we deem the most frequent. This results in 50 word pairs for each language pair.

\begin{table}
\centering
\begin{tabular}{lccS}
\toprule
Lang & GPT-4o & Llama3.1 & {t-stat} \\
\midrule
\texttt{fra} & \textbf{4.79} \small{$\pm$ 0.47} & 4.57 \small{$\pm$ 0.62} & 3.06* \\ 
\texttt{ind} & 4.36 \small{$\pm$ 0.82} & \textbf{4.46} \small{$\pm$ 0.79} & -1.04 \\ 
\texttt{tdt} & \textbf{3.86} \small{$\pm$ 1.18} & 3.61 \small{$\pm$ 1.22} & 1.55 \\
\bottomrule
\end{tabular}
\caption{Average overall rating ($\pm$ standard deviation) for LLM-generated examples per language, with paired t-test results, where * represents a statistically significant difference between models ($p < 0.05$). For rating per criteria, see the distribution bar plot in Figure~\ref{fig:metric-distribution}.}
\label{tab:llm-rating}
\end{table}

\paragraph{Example generation}
We work with two LLMs, GPT-4o \cite{openai2024gpt4technicalreport} and Llama 3.1 405b \cite{dubey2024llama}. The former is the highest rated model overall on the Chatbot Arena as of September 2024~\citep{chiang2024chatbotarenaopenplatform}, the latter is the highest rated among open weights models. For generating example sentence pairs, we use the OpenAI API\footnote{\href{https://platform.openai.com/}{https://platform.openai.com/}} for GPT-4o, and the Replicate API\footnote{\href{https://replicate.com/}{https://replicate.com/}} for Llama 3.1 405b, using a prompt that describes the GDEX criteria and includes the word pairs, shown in Appendix~\ref{sec:prompt-gen-examples}. Both the source and target side sentences are generated jointly in the same output.

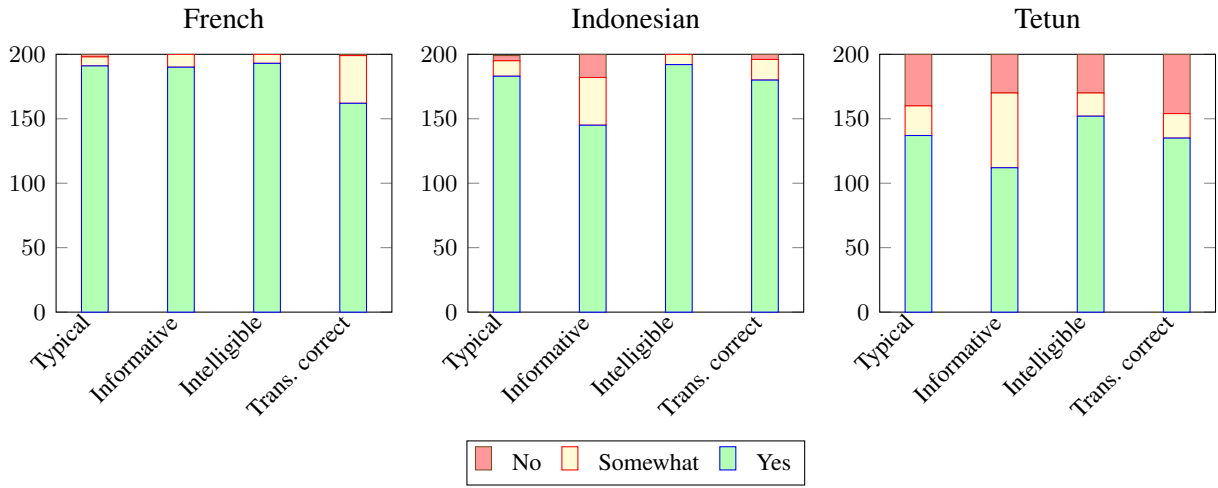
\begin{figure*}
\centering
\begin{tikzpicture}
    \begin{groupplot}[
        group style={
            group size=4 by 1,
            horizontal sep=1cm
        },
        ymin=0, ymax=200,
        ybar stacked,
        yticklabel style={font=\small},
        enlarge x limits=0.15,
        xtick=data,
        xticklabel style={rotate=45, anchor=east, font=\small},
        symbolic x coords={Typical, Informative, Intelligible, Trans. correct},
        width=6cm,
        height=5cm,
        reverse legend
    ]

    \nextgroupplot[title=French, legend to name=sharedlegend, legend columns=3, legend style={column sep=0.5em, font=\small}]
    \addplot+[ybar, fill=green!30] plot coordinates {(Typical,191) (Informative,190) (Intelligible,193) (Trans. correct,162)};\addlegendentry{Yes};
    \addplot+[ybar, fill=yellow!20] plot coordinates {(Typical,7) (Informative,10) (Intelligible,7) (Trans. correct,37)};\addlegendentry{Somewhat};
    \addplot+[ybar, fill=red!40] plot coordinates {(Typical,2) (Informative,0) (Intelligible,0) (Trans. correct,1)};\addlegendentry{No};

    \nextgroupplot[title=Indonesian]
    \addplot+[ybar, fill=green!30] plot coordinates {(Typical,183) (Informative,145) (Intelligible,192) (Trans. correct,180)};
    \addplot+[ybar, fill=yellow!20] plot coordinates {(Typical,12) (Informative,37) (Intelligible,8) (Trans. correct,16)};
    \addplot+[ybar, fill=red!40] plot coordinates {(Typical,4) (Informative,18) (Intelligible,0) (Trans. correct,4)};

    \nextgroupplot[title=Tetun]
    \addplot+[ybar, fill=green!30] plot coordinates {(Typical,137) (Informative,112) (Intelligible,152) (Trans. correct,135)};
    \addplot+[ybar, fill=yellow!20] plot coordinates {(Typical,23) (Informative,58) (Intelligible,18) (Trans. correct,19)};
    \addplot+[ybar, fill=red!40] plot coordinates {(Typical,40) (Informative,30) (Intelligible,30) (Trans. correct,46)};

    \end{groupplot}

    \path (group c1r1.east) -- (group c2r1.west) node[midway] {};
    \node at ($(group c1r1.south)!0.5!(group c3r1.south) - (0,2.0cm)$) {\pgfplotslegendfromname{sharedlegend}};

\end{tikzpicture}
\caption{Rating distributions (GPT-4o and Llama 3.1 combined) for GDEX criteria and translation correctness.}
\label{fig:metric-distribution}
\end{figure*}

\paragraph{Annotator selection and training}
All annotators are native speakers of the source language they rate, and are advanced speakers of English as a second language. We recruit two annotators per source language, one with a computational linguistics background, and one with no background in linguistics or NLP, to get a broad representation of diverse preferences and expectations. Before annotation, we present the task to each annotator, with for each criterion, an explanation of its meaning, along with an example of a sentence that would be rated "Yes" for this criterion, and an example of a sentence that would be rated "No". We explain to each annotator that the "Overall rating" is left to express their general feeling about example quality.

\paragraph{Annotation}
We ask annotators to rate the generated examples against the GDEX criteria (typical, informative, intelligible), with three options for each criterion: ``Yes'', ``Somewhat'', ``No''. After initial observations (on French) that generated sentences can have translation errors, we add another column "Translation correct", with the same options. We also include an "Overall rating" column, where annotators are asked to give their overall impression of the example pair quality, on a scale of 1 to 5 (1 - Bad, 2 - Pretty bad, 3 - Average, 4 - Good, 5 - Very good).


\subsection{Quality of LLM-generated examples}
\label{sec:rating-analysis}

Table~\ref{tab:example-sentences} shows an example of LLM-generated sentences for each language pair, with their associated ratings.

\paragraph{Per language}
Mean overall ratings and annotation distribution are presented in Table~\ref{tab:llm-rating} and Figure~\ref{fig:metric-distribution} respectively. LLM-generated examples get a medium to high overall rating across language pairs. However, there is a clear drop in quality when language is less-resourced. French examples, representing a high-resource language, received the highest ratings (mean 4.68 out of 5), followed by Indonesian (mid-resource, mean 4.41), and then Tetun (low-resource, mean 3.74). This pattern is consistent with previously observed LLM performance degradation on lower-resourced languages \cite{li2024quantifyingmultilingualperformancelarge}, likely due to the reduced amount of training data available for these languages. For example, the MADALAD-400 corpus \cite{kudugunta_madlad-400_2023}, which has documents from Common Crawl tagged by language, has almost 6 times more French documents ($\sim$220M) than Indonesian documents ($\sim$38M), and over 5,000 times more French documents than Tetun documents ($\sim$40k).

\looseness=-1
\paragraph{Per LLM}
Comparing overall rating for the two LLMs used in the study, we find that GPT-4o outperforms Llama3.1 for French (4.79 vs. 4.57), with a statistically significant t-statistic of over 3 indicating a substantial difference between the two models relative to variation in the data. For Indonesian and Tetun however, the paired t-test indicated that the difference between the two models is not statistically significant compared to the variation in the data. We therefore observe variability in LLM output quality that is uneven across languages depending on resource level and shows that performance degradation is not always predictable from resource level.\looseness=-1
\begin{table}
\sisetup{detect-all=true, table-format=2.3}
\centering
\begin{tabular}{lccS}
\toprule
Lang & {A1} & {A2} & {t-stat} \\
\midrule
\texttt{fra} & 4.74 \small{$\pm$ 0.56} & 4.62 \small{$\pm$ 0.56} & 1.830 \\
\midrule
\texttt{ind} & 4.09 \small{$\pm$ 0.85} & 4.73 \small{$\pm$ 0.62} & -6.273* \\
\midrule
\texttt{tdt} & 3.62 \small{$\pm$ 1.47} & 3.85 \small{$\pm$ 0.88} & -1.909 \\
\bottomrule
\end{tabular}
\caption{Average rating ($\pm$ standard deviation) per annotator with paired t-test results, where * represents a statistically significant difference between annotators ($p < 0.05$). For each language, A1 is the annotator with a computational linguistics background. }
\label{tab:rating-per-annotator}
\end{table}

\paragraph{Per GDEX criteria}
Comparing qualitative ratings (typical / intelligible / informative / translation correct), we find a consistent degradation across criteria as the resource level of the language decreased (Figure~\ref{fig:metric-distribution}). For example, 95\% of examples are rated as "typical" for French, but this decreased to 92\% for Indonesian and 69\% for Tetun. The trend was particularly pronounced for the "Informative" criterion (\texttt{fra} 95\%, \texttt{ind} 77\%, \texttt{tdt} 56\%), highlighting the challenges LLMs face in maintaining accurate and relevant examples for lower-resourced languages.\looseness=-1
\paragraph{Per annotator qualification level} Table~\ref{tab:rating-per-annotator} shows no significant difference in mean ratings between annotators for French and Tetun relative to variation in the data, when measured through a paired t-test. For Indonesian, however, we observe a significant and large difference in mean ratings between annotators, where A1 (the annotator with a computational linguistics background) gave much lower ratings than A2.

\subsection{A note on inter-annotator agreement}
\label{sec:inter-annotator-agreement}

\begin{table}
\sisetup{detect-all=true, table-format=1.3}
\centering
\small
\begin{tabular}{llS}
\toprule
Lang & Criteria & {Krippendorff's $\alpha$} \\
\midrule
\multirow{5}{*}{\texttt{fra}} 
& Typical & \textbf{0.378} \\
& Informative & -0.047 \\
& Intelligible & 0.264 \\
& Translation correct & 0.136 \\
& Overall rating & 0.136 \\
\midrule
\multirow{5}{*}{\texttt{ind}} 
& Typical & \textbf{0.517} \\
& Informative & -0.269 \\
& Intelligible & -0.036 \\
& Translation correct & -0.093 \\
& Overall rating & -0.093 \\
\midrule
\multirow{5}{*}{\texttt{tdt}} 
& Typical & \textbf{0.548} \\
& Informative & \textbf{0.449} \\
& Intelligible & \textbf{0.519} \\
& Translation correct & \textbf{0.529} \\
& Overall rating & \textbf{0.529} \\
\bottomrule
\end{tabular}
\caption{Inter-annotator agreement measured using Krippendorff's alpha for different GDEX criteria and overall rating. Bold indicates $\alpha > 0.35$.}
\label{tab:annotator-alpha}
\end{table}

Table~\ref{tab:annotator-alpha} shows relatively low rates of inter-annotator agreement for French and Indonesian, measured through Krippendorff's alpha \cite{castro-2017-fast-krippendorff}, both for overall rating (where individual judgement is encouraged) and for qualitative GDEX criteria (where standard rating is encouraged). For Tetun, however, we observe relatively high inter-annotator agreement across all criteria, including overall rating. We hypothesise that this is due to the more pronounced mistakes in Tetun sentences, which means both that ratings rely less on subtlety of judgement, and that there is more signal to measure. For example, in French, all GDEX criteria are rated "Yes" in over 95\% of examples, giving little room to measure disagreement.

We note that low inter-annotator agreement for rating examples was observed in previous studies \cite{benedetti_automatically_2024}. This finding guides our further experiments: (1) when working with in-context learning, we favour aligning LLM rating with one annotator's judgement at a time, rather than aligning with contradicting ratings coming from multiple annotators (Section~\ref{sec:rating-icl}); (2) when working with pre-trained language models, which are not fine-tuned to annotator preference, we only measure alignment with the annotator who has a computational linguistics background (Section~\ref{sec:predict-gdex}).

\section{Automated rating of example sentences}

Beyond baseline performance across different resource levels, we evaluate how well LLMs can assess example quality. This could enable more efficient dictionary creation pipelines, where automated rating systems that align with human judgement could help filter and select the best examples from larger sets of generated candidates, reducing the need for extensive manual review. Furthermore, reliable automated evaluation metrics could facilitate the development of self-improvement systems where LLMs learn from their own assessments to generate increasingly better examples.

\subsection{Rating through LLM in-context learning}
\label{sec:rating-icl}

For each annotator, we study whether in-context learning can successfully teach the annotator's preferences to an LLM, measured through alignment in overall rating (1-5 score).

\paragraph{Data preparation and model choice}
Given 100 annotated example sentence pairs from a specific annotator, we randomly sample 10 pairs as in-context examples and 90 pairs for evaluation. To avoid bias linked to model self-preference \cite{panickssery2024llmevaluatorsrecognizefavor}, we choose against working with one of the two LLMs used for generating sentences and instead rely on Gemini 1.5 Pro \cite{geminiteam2024gemini15unlockingmultimodal} for this task, given that it is the second best ranked model for instruction following on the Chatbot Arena\footnote{\href{https://lmarena.ai/}{https://lmarena.ai/}} as of September 2024.

\paragraph{Preprocessing through reasoning generation}
For each sentence pair in the sample of 10 pairs, we first ask the LLM to reason about what led to the annotator's rating, given their comment (if any), their ratings of the GDEX criteria, and the translation correctness. Our prompt for this task is provided in Appendix~\ref{sec:reason-rating}.

\paragraph{Evaluation}
We then construct a system prompt that has a list of 10 examples, each with a word and example sentence pair, a reasoning, and final rating from 1 to 5. These examples are injected in the prompt, along with a description of the GDEX criteria (Appendix~\ref{sec:icl-rating}). We use this prompt to ask for a rating for the evaluation of  example pairs.

\begin{table}
\centering
\begin{tabular}{lcc}
\toprule
Lang & Annotator & Rating correl. \\
\midrule
\multirow{2}{*}{\texttt{fra}} & A1-fra & 0.54 \\
& A2-fra & 0.38 \\
\midrule
\multirow{2}{*}{\texttt{ind}} & A1-ind & 0.33 \\
& A2-ind & 0.29 \\
\midrule
\multirow{2}{*}{\texttt{tdt}} & A1-tdt & 0.39 \\
& A2-tdt & 0.42 \\
\bottomrule
\end{tabular}
\caption{Correlation between LLM predicted rating and annotator reference rating (both 1-5) with 10 in-context examples of the annotator's ratings. All correlations are statistically significant with $p < 0.02$.}
\label{tab:rating-pred}
\end{table}

\paragraph{Results}
Table~\ref{tab:rating-pred} demonstrates that in-context learning successfully teaches LLMs annotator preferences across all participants, yielding moderate but significant correlations ranging from 0.29 (A2-ind) to 0.54 (A1-fra). These results span languages of varying resource levels and annotators with diverse backgrounds, highlighting the potential of in-context learning to address challenges related to inter-annotator agreement.

\subsection{Rating through pre-trained language models}
\label{sec:predict-gdex}

In this section, we aim to determine if computationally derived metrics can effectively approximate human judgements of example sentence quality along GDEX criteria.

\paragraph{Data preparation}
We work exclusively with ratings from annotators who have a background in computational linguistics. We map each rating to a number between 0 and 1, where No = 0, Somewhat = 0.5, Yes = 1, allowing us to represent the gradations in quality along a continuous scale.

\paragraph{Metrics and hopythesis}
For each source-side sentence, we compute several metrics using pre-trained language models to test various hypotheses. We examine whether the probability of the entry word (when masked) can serve as a predictor of the "Informative" rating, hypothesising that a lower probability might indicate a more informative context. We also investigate if sentence perplexity can be a good predictor of both the "Intelligible" and "Typical" ratings, with the assumption that lower perplexity could indicate a more intelligible and typical sentence. Additionally, we explore whether context entropy at the position of the entry word could be another predictor of the "Informative" rating, positing that higher entropy might suggest a more informative context.

\paragraph{Choice of models}
To test the hypotheses, we use pre-trained encoder-only language models: \texttt{CamemBERT-large} for French~\cite{martin2019camembert}, \texttt{IndoBERT} for Indonesian~\cite{koto2020indolem}. For Tetun, given the absence of existing encoder-only models for the language, we fine-tune \texttt{XLM-RoBERTa-large}~\cite{xlm-roberta-large} on MADLAD-400~\cite{kudugunta_madlad-400_2023} which is the largest Tetun monolingual corpus available, using the hyperparameters in \citet{adelani_masakhaner_2021}. We release the weights of this model for future researchers.\footnote{\href{https://huggingface.co/raphaelmerx/xlm-roberta-large-tetun}{https://huggingface.co/raphaelmerx/xlm-roberta-large-tetun}}

\paragraph{Results}
As Table~\ref{tab:mlm-correlation} demonstrates, the probability of the target word serves as a fair predictor of informativeness for French, with a correlation of 0.21, but this relationship does not hold for other languages. High perplexity proves to be a moderately good predictor of low intelligibility for both French and Indonesian, with correlations of -0.57 and -0.52 respectively. Similarly, high perplexity is a good predictor of low typicality for French (correlation of -0.41) and moderately good for Indonesian (-0.32). Notably, no significant correlations are found for Tetun across these metrics. Contrary to our hypothesis, context entropy at the target word (when masked) does not serve as a good predictor for informativeness across any of the languages studied.

\begin{table}
\centering
\begin{tabular}{lll S}
\toprule
Lang & Criterion & LM Metric & {Correl.} \\
\midrule
\multirow{4}{*}{\texttt{fra}} & Informative & Word Prob. &  0.210* \\
& Intelligible & Perplexity & \bfseries -0.566* \\
& Typical & Perplexity & \bfseries -0.408* \\
& Informative & Entropy &  0.062 \\
\midrule
\multirow{4}{*}{\texttt{ind}} & Informative & Word Prob. & 0.176 \\
& Intelligible & Perplexity & \bfseries -0.521* \\
& Typical & Perplexity & \bfseries -0.320* \\
& Informative & Entropy & 0.124 \\
\midrule
\multirow{4}{*}{\texttt{tdt}} & Informative & Word Prob. & 0.113 \\
& Intelligible & Perplexity & 0.101 \\
& Typical & Perplexity & 0.136 \\
& Informative & Entropy &  0.068 \\
\bottomrule
\end{tabular}
\caption{Correlation between GDEX ratings and masked LM metrics. * denotes statistical significant with $p < 0.05$.}
\label{tab:mlm-correlation}
\end{table}

\paragraph{Implications}
Our results show the potential of sentence perplexity for estimating example sentence typicality and intelligibility, for middle- to high-resource languages. The lack of significant results for Tetun demonstrates that the amount of available corpora in this low-resource language is not sufficient to get a pre-trained language model that captures sentence quality with a high degree of accuracy.

\section{Discussion}
\label{sec:discussion}

Our study provides several insights into the capabilities and limitations of LLMs for generating and evaluating bilingual dictionary examples. First, we demonstrate that LLMs are capable of producing reasonably good quality example sentences across multiple language pairs. However, there is a clear degradation in performance as we move from high-resource languages like French to low-resource languages like Tetun. The variability in output quality across languages underscores the need for careful evaluation and potential supplementary techniques when applying LLMs to lexicographic tasks, especially for less-represented languages.

A notable challenge revealed in our study is the high variance in personal preferences for example sentence quality, as evidenced by low inter-annotator agreement rates. This variability poses difficulties in establishing a single, universally accepted metric for evaluating dictionary examples. However, our experiments with in-context learning demonstrate that LLMs can be successfully aligned with individual annotator preferences, even for low-resource languages like Tetun. This finding suggests a promising avenue for tailoring LLM outputs to specific lexicographic standards or individual annotator judgements, potentially facilitating the example generation and evaluation process.

The low inter-annotator agreement observed in our study highlights the need for annotations from multiple annotators before drawing conclusions about the quality (or lack thereof) of example sentences. This multi-annotator approach can help capture a more comprehensive range of perspectives and mitigate individual biases. Additionally, our findings, particularly for French where most GDEX criteria were rated "Yes" due to the high quality of generated sentences, suggest the need for finer measures of criteria to better capture nuanced levels of quality. We recommend developing more granular rating scales or additional sub-criteria, especially for high-resource languages where LLMs perform well. This refinement in evaluation methods could provide more discriminative assessments of LLM-generated example sentences.

\section{Conclusion}

We contribute a first evaluation of LLM capability to generate bilingual example sentences, across languages of various resource levels. We show that although LLMs are capable of generating good bilingual example sentences on average, their performance degrades with language resource level. We further show that even when using a shared framework for sentence evaluation (GDEX), annotators tend to disagree with each other on sentence quality, but that in-context learning can be leveraged to align LLMs with a specific annotator's ratings.

Our findings highlight the potential of LLMs in lowering the cost of lexicographic work, and their ability in aligning with human judgement in a field where human judgement can be highly variable. This is of particular value in low-resource lexicographic work, where lack of human resources may prevent the widespread compilation of lexicographic resources.

\section*{Limitations}

While our study shows LLMs can play a helpful role in the generation and rating of bilingual dictionary examples, our choice of experiment constraints can limit the reach of our results. We work exclusively with languages that use Latin script, and with English on the target side, which raises the question of how our results would hold for languages that use other scripts and with lower-resource target languages. We did not include part of speech information when generating examples, and do not study performance on words that have several definitions; both choices may have skewed the quality of generated example downwards.

The low inter-annotator agreement, while part of the experiment, and expected in this lexicographic context, raises questions about how we could have better aligned annotators, for example by using pre-qualifying questions, or by exclusively relying on linguists for annotation.

We identify several areas for future work. First, LLM rating of example sentences could be integrated in the example generation pipeline, for instance by having an LLM generate a number of candidate examples, and another LLM automatically rank them, similar to the approach by \citet{cai_low-cost_2024}. Second, the quality of LLM-generated example sentences could be compared against sentences retrieved from a corpus. Last, the incorporation of retrieved sentences in the LLM prompt could guide the LLM to generate more typical or informative sentences.

\section*{Acknowledgments}
We would like to extend our gratitude to Professor Hanna Suominen for her valuable feedback and guidance throughout this study. We also thank Gabriel de Jesus and his team, Isabel Pereira~(Catalpa International), Tungga Dewi, and Matilda Merx for their support in data collection and annotation. We also thank the anonymous reviewers for their feedback. This research was supported by The University of Melbourne’s Research Computing Services and the Petascale Campus Initiative.


\bibliography{custom, references_zotero}

\appendix

\section{Prompts used}
\label{sec:prompts}

In the prompts below, the parts in brackets (e.g. \verb|{SRC_NAME}|) are templated out.

\subsection{Generating examples}
\label{sec:prompt-gen-examples}

\begin{lstlisting}

You are assisting in the creation of a bilingual {SRC_NAME}-{TGT_NAME} dictionary. Your task is to generate example sentences for dictionary entries to help users understand the usage of words in context.

You will be provided with a {SRC_NAME} word and its {TGT_NAME} equivalent.
<{SRC_NAME} entry>
{{src_word}}
</{SRC_NAME} entry>

<{TGT_NAME} entry>
{{tgt_word}}
</{TGT_NAME} entry>

Please create a pair of example sentences for each entry. The sentences should be:
1. Typical: Show typical usage of the word
2. Informative: Add value by providing context or additional information
3. Intelligible: Be clear, concise, and appropriate for a general audience
4. Using the entries provided above (the {SRC_NAME} and {TGT_NAME} words)

Format your response as follows:

<example_sentence_pair>
{SRC_NAME}: [Your {SRC_NAME} sentence here]
{TGT_NAME}: [Your {TGT_NAME} sentence here]
</example_sentence_pair>

Please provide your example sentences based on the given {SRC_NAME} and {TGT_NAME} entries.
\end{lstlisting}

\subsection{Reasoning about a specific annotator's rating}
\label{sec:reason-rating}

\begin{lstlisting}
<example>
Src Entry: {src_entry}
Tgt Entry: {tgt_entry}
Src Example: {src_example}
Tgt Example: {tgt_example}

Comment: {comment}
Typical: {typical}
Informative: {informative}
Intelligible: {intelligible}
Translation correct: {translation_correct}
</example>

Reasoning: what is the reasoning for the above ratings? Give your response in one paragraph.
\end{lstlisting}

\subsection{In-context learning for aligning an LLM with an annotator}
\label{sec:icl-rating}

\subsubsection{Prompt construction}

\begin{lstlisting}[language=Python]
TEMPLATE_EXAMPLE = """<example>
<data>
Src Entry: {src_entry}
Tgt Entry: {tgt_entry}
Src Example: {src_example}
Tgt Example: {tgt_example}
</data>
<reasoning>{reasoning}</reasoning>
<rating>{rating}</rating>
</example>"""

def get_templated_example(row):
    return TEMPLATE_EXAMPLE.format(
        src_entry=row[SRC_LANG],
        tgt_entry=row[TGT_LANG],
        src_example=row['src_example'],
        tgt_example=row['tgt_example'],
        reasoning=row['reasoning'],
        rating=row['Overall rating']
    )

AUGMENTED_SYSTEM_PROMPT = SYSTEM
for row in sample:
    AUGMENTED_SYSTEM_PROMPT += get_templated_example(row)
    AUGMENTED_SYSTEM_PROMPT += '\n\n'
\end{lstlisting}

\subsubsection{Prompt example}

An example constructed prompt with two examples. Note that our experiments used 10 examples.

\begin{lstlisting}
You are assisting in the creation of a bilingual English-Indonesian dictionary.
Your task is to rate a candidate sentence pair that illustrates dictionary entries to help linguists select an appropriate example pair.

Example sentences should should be:
1. Typical: Show typical usage of the word
2. Informative: Add value by providing context or additional information
3. Intelligible: Be clear, concise, and appropriate for a general audience
4. Translation correct: Are sentences a good translation of each other, with fluent grammar and correct usage of words in both languages

You are rating the example sentences, not the dictionary entries.

<example>
<data>
Src Entry: meriam
Tgt Entry: cannon
Src Example: Meriam itu ditempatkan di atas bukit untuk melindungi kota dari serangan musuh.
Tgt Example: The cannon was placed on the hill to protect the city from enemy attacks.
</data>
<reasoning>The example sentences are typical because they demonstrate a standard use of the word "cannon" in a military context.  However, they are only somewhat informative because the statement about cannons being used for defense, while not entirely inaccurate, might not be the most common understanding. The sentences are intelligible due to their clear and concise language, and the translation is accurate, reflecting the meaning and grammar of both the source and target languages. 
</reasoning>
<rating>4 Good</rating>
</example>

<example>
<data>
Src Entry: menanyai
Tgt Entry: question
Src Example: Polisi menanyai saksi mata untuk memperoleh informasi lebih lanjut tentang kejadian itu.
Tgt Example: The police questioned the eyewitness to obtain more information about the incident.
</data>
<reasoning>The ratings are justified because the sentences demonstrate typical usage of the words "menanyai" and "questioned" in the context of a police investigation. They are informative by providing context about the purpose of the questioning. Both sentences are clear and concise, making them intelligible. However, the translation is slightly off because "keterangan" would be a more natural choice than "informasi" in Indonesian, making the translation somewhat less accurate. 
</reasoning>
<rating>4 Good</rating>
</example>

...

<data>
Src Entry: sehari-hari
Tgt Entry: everyday
Src Example: Saya menggunakan sepeda sebagai alat transportasi sehari-hari karena lebih ramah lingkungan.
Tgt Example: I use a bicycle as my everyday mode of transportation because it's more environmentally friendly.
</data>
\end{lstlisting}

\end{document}